\documentclass{sig-alternate}

\usepackage{algorithmic,algorithm}

\newcommand{\Real}{\mathop{\rm I\kern-.2emR}}
\newcommand{\V}{{\cal N}}

\begin{document}
\conferenceinfo{GECCO'11,} {July 12--16, 2011, Dublin, Ireland.}
\CopyrightYear{2011}
\crdata{978-1-4503-0557-0/11/07}
\clubpenalty=10000
\widowpenalty = 10000

\title{
The Road to VEGAS:\\
Guiding the Search over Neutral Networks
}
%\subtitle{[Extended Abstract]
%\titlenote{A full version of this paper is available as
%\textit{Author's Guide to Preparing ACM SIG Proceedings Using
%\LaTeX$2_\epsilon$\ and BibTeX} at
%\texttt{www.acm.org/eaddress.htm}}}
%
% You need the command \numberofauthors to handle the 'placement
% and alignment' of the authors beneath the title.
%
% For aesthetic reasons, we recommend 'three authors at a time'
% i.e. three 'name/affiliation blocks' be placed beneath the title.
%
% NOTE: You are NOT restricted in how many 'rows' of
% "name/affiliations" may appear. We just ask that you restrict
% the number of 'columns' to three.
%
% Because of the available 'opening page real-estate'
% we ask you to refrain from putting more than six authors
% (two rows with three columns) beneath the article title.
% More than six makes the first-page appear very cluttered indeed.
%
% Use the \alignauthor commands to handle the names
% and affiliations for an 'aesthetic maximum' of six authors.
% Add names, affiliations, addresses for
% the seventh etc. author(s) as the argument for the
% \additionalauthors command.
% These 'additional authors' will be output/set for you
% without further effort on your part as the last section in
% the body of your article BEFORE References or any Appendices.

\numberofauthors{5} %  in this sample file, there are a *total*
% of EIGHT authors. SIX appear on the 'first-page' (for formatting
% reasons) and the remaining two appear in the \additionalauthors section.
%
\author{
% You can go ahead and credit any number of authors here,
% e.g. one 'row of three' or two rows (consisting of one row of three
% and a second row of one, two or three).
%
% The command \alignauthor (no curly braces needed) should
% precede each author name, affiliation/snail-mail address and
% e-mail address. Additionally, tag each line of
% affiliation/address with \affaddr, and tag the
% e-mail address with \email.
%
% 1st. author
\alignauthor Marie-\'El\'eonore Marmion
       \affaddr{Universit\'e Lille 1}\\
       \affaddr{LIFL -- CNRS -- INRIA}\\
       \email{marie-eleonore.marmion@lifl.fr}
% 2nd. author
\alignauthor Clarisse Dhaenens
       \affaddr{Universit\'e Lille 1}\\
       \affaddr{LIFL -- CNRS -- INRIA}\\
       \email{clarisse.dhaenens@lifl.fr}
% 3rd. author
\alignauthor Laetitia Jourdan\\
       \affaddr{Universit\'e Lille 1}\\
       \affaddr{LIFL -- CNRS -- INRIA}\\
       \email{laetitia.jourdan@inria.fr}
%\more-auths
\and
% 4th. author
\alignauthor Arnaud Liefooghe
       \affaddr{Universit\'e Lille 1}\\
       \affaddr{LIFL -- CNRS -- INRIA}\\
       \email{arnaud.liefooghe@lifl.fr}
% 5th. author
\alignauthor S\'ebastien Verel\\
       \affaddr{Univ. Nice Sophia-Antipolis}\\
       \affaddr{INRIA}\\
       \email{sebastien.verel@inria.fr}
}
% There's nothing stopping you putting the seventh, eighth, etc.
% author on the opening page (as the 'third row') but we ask,
% for aesthetic reasons that you place these 'additional authors'
% in the \additional authors block, viz.
%\additionalauthors{Additional authors: John Smith (The Th{\o}rv{\"a}ld Group,
%email: {\texttt{jsmith@affiliation.org}}) and Julius P.~Kumquat
%(The Kumquat Consortium, email: {\texttt{jpkumquat@consortium.net}}).}
\date{April 5th 2011}
% Just remember to make sure that the TOTAL number of authors
% is the number that will appear on the first page PLUS the
% number that will appear in the \additionalauthors section.

\maketitle
\begin{abstract}
VEGAS (Varying Evolvability-Guided Adaptive Search) is a new methodology proposed to deal with
the neutrality property that frequently appears on combinatorial optimization problems. 
Its main feature is to consider the whole evaluated solutions of a neutral network rather than the last accepted solution.
Moreover, VEGAS is designed to escape from plateaus based on the evolvability of solutions,
and on a multi-armed bandit by selecting the more promising solution from the neutral network.
Experiments are conducted on NK-landscapes with neutrality.
Results show the importance of considering the whole identified solutions from the neutral network and of guiding the search explicitly.
The impact of the level of neutrality and of the exploration-exploitation trade-off are deeply analyzed.
\end{abstract}




%============================================================================
\section{Motivations}

Due to their ability to find satisfying solutions with high efficiency and effectiveness,
the design of local search approaches is still a prominent issue for hard combinatorial optimization.
This class of methods is also of strong interest since they are generally quite simple to implement.
However, in some circumstances, these methods may have more difficulties.
One of the most critical situation arises when the problem under study holds neutrality,
which is the case for many problems from combinatorial optimization, like scheduling or satisfiability.
This property means that a lot of different solutions share the same fitness value.
In such a case, natural questions hold:
Once the neutrality of a problem is known, how could the search exploit it?
How could the search be guided to exploit this neutrality with success?
The aim of this article is to propose an efficient approach, based on simple local search principles and adaptive mechanisms,
that exploits the inherent neutrality to a problem, without inhibiting too much the search for neutrality-free problems.
Therefore, an experimental comparison will be driven between approaches
that do or do not exploit the neutrality explicitly during the search process.

The central idea of the article is based on two observations.
On the one hand, up to now, the neutrality property has been under-exploited on the design of search algorithms,
with the exceptions of a few examples that will be detailed later in the paper.
On the other hand, adaptive search, based on a multi-armed bandit framework, has been successfully applied
to parameter control in the context of evolutionary algorithms, in particular in adaptive operator selection \cite{Fialho10}.
In this paper, the goal is not to adapt parameters along the search.
Instead, we propose an original adaptive algorithm, based on a multi-armed bandit framework,
to guide the search over neutral networks.
When the search is stuck on a plateau, VEGAS adaptively selects the more-promising solution whose neighborhood has to be explored.
Hence, two main questions are addressed in this paper: 
\begin{itemize}
\item[($i$)] How to consider neutrality, and then neutral networks, during the search process?
How to avoid an arbitrary choice to be made during a neutral search?
\item[($ii$)] How to guide the search adaptively over neutral networks
in order to explore the neighborhood of a solution that is more likely to escape a plateau by the top?
\end{itemize}
The paper is organized as follows.
Section \ref{sec:background} introduces fundamental definitions
together with previous works on neutrality-based and adaptive search methods.
The proposed VEGAS algorithm, where the search is adaptively guided by the evolvability of solutions, is introduced in Section \ref{sec:vegas}.
Section~\ref{sec:exp} presents experimental results on the performance of VEGAS,
on the influence of its single problem-independent parameter and on the influence of the degree of neutrality of the problem at hand.
The last section gives a summary of results and discusses future works.

%============================================================================
\section{Background}
\label{sec:background}

This section introduces the main concepts dealing with landscape analysis and the design of neutrality-based and adaptive search methods.

\subsection{Local Search, Neutrality and Evolvability}
\label{subsec:nbh}

A fitness landscape \cite{stadler:1995} is a triplet $(S, \V, f)$ where $S$ is a set of admissible solutions ({\itshape i.e.} the search space),
$\V: S \longrightarrow 2^{S}$ is a neighborhood structure, 
and $f: S \longrightarrow \Real$ is a fitness function that can be pictured as the \textit{height} of the corresponding solutions, here assumed to be maximized.
A \emph{neighborhood structure} is a mapping function
that assigns a set of solutions $\mathcal{N}(s) \subset S$ to any feasible solution $s \in S$.
$\mathcal{N}(s)$ is called the \emph{neighborhood} of $s$, 
and a solution $s' \in \mathcal{N}(s)$ is called a \emph{neighbor} of $s$.
We then may define a \emph{local optimum} as: solution $s^{*}$ is a \emph{local optimum} iff no neighbor has a better fitness value:
$\forall s \in \V(s^{*})$, $f(s^{*}) \geq f(s)$. 

The importance of {\em selective neutrality} as a significant factor in evolution was stressed by Kimura \cite{Kimura1983} in the context of evolutionary theory. 
The relevance and benefits of neutrality for the robustness and evolvability in living systems have been recently discussed by Wagner \cite{Wagner2005}.
There is a growing evidence that such large-scale neutrality is also present in artificial landscapes. Not only in combinatorial fitness landscapes such as randomly generated SAT instances~\cite{frank-et-al97},  cellular automata rules~\cite{verel07} and many others,  but also in complex real-world design and engineering applications such as evolutionary robotics~\cite{husbands,Smith2002}, evolvable hardware~\cite{HarveyT96,Vassilev2000},  genetic programming~\cite{Banzhaf1994,yu01} and scheduling \cite{Marmion:Lion2011}. 

A  \emph{neutral neighbor} of a given solution $s \in S$ is a neighboring solution $s^\prime \in \V(s)$ with the same fitness value:
$f(s^\prime) = f(s)$.
Given a solution $s \in S$, the set of neutral solutions in its neighborhood is defined by $\V_n(s) = \{ s' \in \V(s) ~|~ f(s') = f(s) \}$.
The \emph{neutral degree} of a solution is the number of its neutral neighbors.
A fitness landscape is said to be \emph{neutral} if there are many solutions with a high neutral degree. 

A  \emph{neutral network}, or \emph{plateau}, denoted here by NN, is a connected sub-graph of solutions with respect to neighborhood relation whose vertices are solutions with the same fitness value. There exists an edge between solutions $s$ and $s^\prime$ when $s^\prime$ is a neutral neighbor of $s$. 
%Two vertices in a NN are connected if they are neutral neighbors:
%$ \forall s \in \mbox{NN}, \exists s' \in \mbox{NN}$ such that $s \in \V(s')$ and $f(s^\prime) = f(s)$.
A \textit{portal} in a NN is a solution that has at least one neighbor with a better fitness value, {\itshape i.e.} a greater fitness value in a maximization context.

%-------------------------------------
\subsection{Neutrality-based Search}

When dealing with neutrality, two extreme local search approaches are usually designed.
The first one simply ignores neutrality. 
One of the simplest algorithm is the first-improvement hill-climbing (FIHC),
where the first evaluated neighbor that \textit{strictly} improves the current fitness value is accepted.
In other words, the heuristic does not move on a neutral neighboring solution,
and prefers to keep exploring the neighborhood of the current solution,
assuming that the neutral neighbor will not lead to better solutions.
At the opposite, some local search approaches always accept the first visited neighbor with the same fitness value. 
A typical example is the \emph{Netcrawler} process \cite{barnett01},
that consists of a random neutral walk with a mutation mode adapted to local neutrality.
The per-sequence mutation rate is optimized to jump from one NN to another.
Stewart \cite{stewart2001} also proposed an \emph{Extrema Selection} for evolutionary optimization
in order to move on promising solutions in a neutral search space. 
The selection aims at accelerating the evolution during the search process once most solutions from the population
have reached the same level of performance.
To each solution is assigned an endogenous performance during the selection step to explore the search space area with the same performance more largely, with the assumption that it will help to reach solutions with better fitness values.
Additionally, the NILS (Neutrality-based Iterated Local Search) algorithm, recently proposed in \cite{Marmion:Evo2011},
shows interesting results and enhances the interest of taking neutrality into account for flowshop scheduling problems.

All those heuristics focus the search on the last solution found along the NN.
Their bet is that the new accepted solution has more chance to lead to better solutions than the previous one,
because no better solution was (yet) evaluated in its neighborhood.
But, when no better neighbor has been found again, 
the heuristic prefers to move to a new solution (in another part of the search space) than to go back on a previously accepted solution,
even if it could seem more promising \textit{a posteriori}.
Hence, nothing motivates the choice of exploring the neighborhood of the last-accepted solution from the NN rather than any other.
Most of the time, such a choice appears quite arbitrary.
We believe that there might exist a better trade-off between these two extreme cases
(ignoring neutrality, and starting with the last-accepted neutral solution),
by keeping memory of the solutions evaluated along the NN.
%of all the NN that was previously evaluated.

%-------------------------------------
\subsection{Adaptive Search}
\label{sec:adaptiveSearch}

Autonomous self-management search receives more and more attention from the past years
due to the increasing complexity of the search methods and problems.
The general goal of those methods is to automatically adapt their mechanism to the changing problem conditions.
The aim of parameter control is the on-line setting of parameters 
such as the representation of solution, the stochastic operators (mutation, crossover), 
the selection operators, the application rate of those operators, etc. \cite{EibenMSS07}.
In combinatorial optimization,
adaptive methods are often preferred over self-adaptive ones which increase the size of the search space.
From the search history, adaptive methods select the new parameter setting.
Different rules are used for selection:
probability matching, adaptive pursuit \cite{Thierens:2005} which attaches a probability of success to each operator,
the multi-armed bandit \cite{Fialho08}, and so on.
The multi-armed bandit framework is a sequential learning model, mostly studied in game theory,
dealing with the trade-off between exploration and exploitation.
It considers a set of $K$ independent arms, 
each one having some reward following an unknown distribution.
An optimal selection strategy maximizes the cumulative reward along time.
The Upper Confidence Bound (UCB) strategy \cite{Auer:2002:FAM:599614.599677}, which is asymptotically optimal,
has been used in the context of adaptive operator selection \cite{Fialho08}.
To each arm, which is an operator in that context, 
is associated an empirical reward which reflects its quality.
Then, the operator (arm) with the best score is selected, 
where the upper confidence bound of the reward defines this score:

\begin{equation}
%\arg\max_{i=1..K} \left(
\text{score}_{i,t} = \hat{r}_{i,t}  + C \sqrt{\frac{\log \sum_{k} n_{k,t}}{n_{i,t}}} 
\label{equation1}
\end{equation}
where for the time step $t$, 
$\hat{r}_{i,t}$ is the empirical reward of operator $i$, 
$n_{i,t}$ is the number of times that the operator $i$ has been tried.
$C$ is a problem-independent parameter representing a scaling factor which controls the trade-off between exploitation and exploration.

\begin{figure}
\begin{center}
\includegraphics{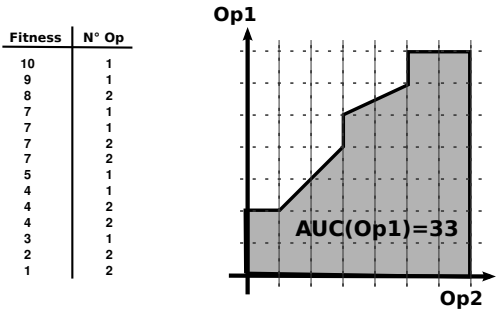}
\caption{
Example of the AUC reward computed for an operator $op_1$.
The list of fitness values produced by each possible operator is sorted in the decreasing order.
A curve from the point $(0 , 0)$ is drawn according to this list: 
When the operator $op_1$ is concerned, 
the curve follows the $op_1$ axe,
otherwise it follows the other axe.
The area under this curve is the reward of the operator $op_1$ \cite{Fialho10}.}
\label{auc}
\end{center}
\end{figure}
The measure of the operator quality (reward) has an impact on the efficiency of the method.
Different measures of reward have been proposed: 
the average fitness improvement between parent and offspring,
the maximum fitness improvement from a time windows \cite{Fialho08}, 
or more recently \textit{the Area Under the Curve} \cite{Fialho10}.
This credit assignment uses the comparison of solution fitness values produced by each operator.
This method uses the rank instead of a normalized fitness improvement.
Then, the bandit adaptive technique becomes more invariant to fitness function transformation,
and the sensitivity according to the parameter $C$ decreases.
The Figure \ref{auc} gives an example of AUC computation.
The details is not given due to space limitation, 
the reader is refereed to~\cite{Fialho10}.
%============================================================================
\section{VEGAS}
\label{sec:vegas}

This section presents the VEGAS (Varying Evolvability-Guided Adaptive Search) algorithm. 
As seen in the previous section, the First-Improvement Hill-Climbing (FIHC) algorithm and the Netcrawler (NC) algorithm
are based on different strategies.
The first one  does not take neutrality into account, whereas the second one proposes a specific way to deal with it (by always exploring the last accepted - better or neutral - solution).
The aim of VEGAS is to take the neutrality explicitly into account and to propose an efficient way to guide the search on NN.
First, we reconsider the search over a NN.
Second, a guiding strategy, based on the evolvability of solutions, is proposed.

\subsection{Considering Neutral Networks}

First, let us define some useful terms. 
A solution is said to be \emph{evaluated} if its fitness value has been computed. 
A solution is marked as \emph{visited} if its neighborhood has been completely evaluated, otherwise it is \emph{non-visited}.
The neighborhood of a solution is explored in a random order without repetition.
Let us remind that a NN, also known as \emph{plateau}, is a set of neighboring solutions with the same fitness value.
The set of evaluated solutions from the current NN is denoted by $S$.

Let us consider a simple local search algorithm that iteratively improves a current solution $s$ by exploring its neighborhood.
As soon as a strictly improving neighboring solution is found, it is accepted and replaces the current solution.
As long as a neutral neighboring solution is evaluated, a particular strategy is applied in order to iteratively build the set $S$.
As opposed to a NC, the main idea of our approach is to consider the whole set of evaluated solutions from the current NN (\textit{i.e.} $S$)
instead of a single one (the last-evaluated solution).
Now, the question is: Which solution $s \in S$ to select in order to evaluate a new neighboring solution?
For instance, when no particular information is computed, this solution can be selected at random.
The algorithm stops once all solutions from $S$ are marked as visited,
\textit{i.e.} the neighborhood of all solutions from $S$ has been explored.
When there is no neutrality, this algorithm behaves like a FIHC.

\subsection{Guiding the Search over Neutral Networks}

% algorithm
\begin{algorithm}[t]
\caption{VEGAS}
\label{algoNNS}
\begin{algorithmic}
\STATE $S=\{s_0\}$
\WHILE{$\exists s \in S$ such that $s$ is not visited}
	\STATE $s \leftarrow$ \it{select($S$)}
	\STATE Choose a solution $s' \in \mathcal{N}(s)$ at random (no repetition)
	\IF{$f(s')>f(s)$}
		\STATE $S \leftarrow \{s'\}$
	\ELSIF{$f(s')=f(s)$}
		\STATE $S \leftarrow S\cup\{s'\}$
	\ENDIF
	\STATE Update rewards($s,s'$)
\ENDWHILE
\STATE \textbf{Return} $s \in S$
\end{algorithmic}
\end{algorithm}

Instead of randomly choosing the next solution to explore, the selection can be guided.
Indeed, on a NN, $|S|>1$ solutions can be explored.
Only one has to be chosen in order to evaluate one of its neighbors.
To do so, we here propose to use the evolvability of solutions. 

Algorithm \ref{algoNNS} presents the general framework of VEGAS.
All the evaluated solutions of the current NN are recorded in $S$.
Then, a \textit{select} method returns a solution $s \in S$.
A new neighbor $s'\in \mathcal{N}(s)$ is evaluated (without repetition).
If $s'$ has a better fitness value, the set $S$ becomes the singleton $\{s'\}$.
Otherwise, if $s'$ has the same fitness value than the current fitness value, it is added to $S$.
A reward is computed for each solution from $S$, and the \textit{select} method is applied. 

The \textit{select} method is one of the main component of VEGAS.
For instance, if \textit{select($S$)} always returns the last neutral solution evaluated, this algorithm behaves like a NC.
Here, we consider that the solution with the highest score, as given by equation (\ref{equation1}), is selected.
Thus, every time a new solution $s^\prime$ is evaluated in the neighborhood of a solution $s \in S$,
$s^\prime$ is recorded to update the score values of all solutions in $S$ according to the credit assignment under consideration.
The reward is based on the AUC (see Section~\ref{sec:adaptiveSearch}).
The arms are the solutions from $S$, 
and the fitness values of the evaluated neighbors are used to compare solutions.
The AUC gives a way to compare the evolvability of evaluated solutions on a NN.
For instance, when the fitness value of neighbors from solution $s \in S$ are better than those of solution $s^\prime \in S$,
the AUC of $s$ is higher than the one of~$s^\prime$.

%In order to compute score values, the evaluated neighboring solutions are ranked according to their respective fitness values.
%The ranks of the evaluated neighbors of a solution $s \in S$ are required to compute the score value as proposed in~\cite{Fialho10}.
%Indeed, the AUC (Area Under the Curve) method computes the first term of the score (\ref{equation1}) of each solution from $S$,
%according to the rank of the corresponding evaluated neighboring solutions.

The parameter $C$ in (\ref{equation1}) allows to tune the trade-off between exploration and exploitation.
Here, it affects the exploration and the exploitation of the neighborhood of solutions from the NN.
When $C$ is large, it gives more weight to exploration: the search promotes the sampling of neighborhoods with few solutions evaluated.
When $C$ is small, it gives more weight to exploitation: the search promotes the sampling of neighborhoods where
the best neighbors have been evaluated so far.
This is based on the assumption that solutions with a higher evolvability are more likely to find a portal.

%============================================================================
\section{Experiments}
\label{sec:exp}
%\input{part4}

%-------------------------------------
%-------------------------------------
\subsection{NK-Landscapes with Neutrality}

The family of $NK$-landscapes is a problem-independent model used for constructing multimodal landscapes \cite{kauffman93}.
Such a model is of high interest in order to design new search approaches.
Parameter $N$ refers to the number of bits in the search space ({\itshape i.e.} the binary string length),
and $K$ to the number of bits that influences a particular bit from the string (the epistatic interactions).
By increasing the value of $K$ from 0 to $(N-1)$, $NK$-landscapes can be gradually tuned from smooth to rugged.
The fitness function (to be maximized) of a $NK$-landscape $f_{NK}: \lbrace 0, 1 \rbrace^{N} \rightarrow [0,1)$ is defined on binary strings of size $N$. An `atom' with a fixed epistasis level is represented by a fitness component $f_i: \lbrace 0, 1 \rbrace^{K+1} \rightarrow [0,1)$,
associated with each bit $i \in N$.
Its value depends on the allele at bit $i$ and also on the alleles at $K$ other epistatic positions ($K$ must be defined between $0$ and $N - 1$).
The fitness $f_{NK}(x)$ of a solution $x \in \lbrace 0, 1 \rbrace^{N}$ corresponds to the mean value of its $N$ fitness components $f_i$: \label{defNK}
$ f_{NK}(x) = \frac{1}{N} \sum_{i=1}^{N} f_i(x_i, x_{i_1}, \ldots, x_{i_K})$,
where $\lbrace i_1, \ldots, i_{K} \rbrace \subset \lbrace 1, \ldots, i-1, i+1, \ldots, N \rbrace$.
In the original $NK$-landscapes, the fitness components are uniformly distributed in the range $[0,1)$,
so that it is very unlikely that the same fitness value is assigned to two different solutions.
In other words, the neutrality is \textit{null}.

In our study, we will use an extension of this initial model in which neutrality has been added.
The way the neutrality is artificially included has an important impact on the structure of the resulting landscapes.
Several models of neutrality have been proposed to generalize the initial $NK$-landscapes by adding a tunable level of neutrality.
Among others, there are the $NK_p$-landscapes ($p$ is for probabilistic) \cite{Barnett98},
and the $NK_q$-landscapes ($q$ is for quantized) \cite{Newman98}. 
$NK_p$-landscapes are very similar to $NK$-landscapes,
except that the fitness contribution are null with the rate $p$.%there is a given probability $p$ that the fitness contribution of a bit is set to $0$.
In the $NK_q$-landscapes, that will be used in the paper, the fitness contributions are integer values belonging to the range $[0, q)$.
The total fitness is scaled by a factor $1 /(q-1)$ in order to translate it in the range $[0, 1]$.
As indicated by Geard et al. \cite{Geard02} in their comparison of neutral landscapes,
$NK_q$-landscapes are similar to the $NK$-landscapes in several aspects.
The $NK_q$-landscapes look like $NK$-landscapes in which rugged hillsides have been flattened into plateaus.
The smaller $q$, the higher the level of neutrality.

%-------------------------------------
%-------------------------------------
\subsection{Experimental Design}

In order to study the performance of the proposed VEGAS algorithm, we compare it with three other approaches:
\begin{itemize}
\item FIHC: a First-Improvement Hill Climbing algorithm, that \textit{strictly} improves the current fitness value during the search;
\item NC: a Netcrawler algorithm \cite{barnett01}, that allows neutral moves to be performed during the search.
%NC behaves like a FIHC, except that the first evaluated neutral neighbor is now accepted.;
\item F2NS: a Fair Neutral Network Search algorithm, that evaluates a random neighbor (without replacement) of a random solution from the NN.
\end{itemize}
We experiment these algorithms on randomly-generated $NK_q$-landscapes with $N=64$,
$K \in \{ 2, 4, 6, 8 \}$ and $q \in \{ 2, 3, 4 \}$. 
They will give us the opportunity to compare these algorithms according to different configurations with respect to neutrality and non-linearity.
The neighborhood is defined with the bit-flip operator of one bit.
For every instance, $100$ independent executions are performed.%each algorithm is run $100$ times.

All the algorithms start with a random solution.
The stopping condition is given in terms of a maximal number of evaluations, set to $10^5$.
The four algorithms could converge before the stopping condition.
Thus, they have all been included in a random-restart framework:
when an algorithm seems to have converged, 
the search restarts with a new random solution (keeping the best ever found solution).
For the FIHC,
the search stops when the current solution is a local optimum (no neighbor has a strictly higher fitness).
For the NC,
the search stops on solutions which are strict local optima where the current can not be strictly improve.
For the NC, we fix a second maximal number of evaluations denoted by $k$.
This number has been set according to $K$ and $q$ from preliminary experiments.
For every instance, $30$ independent runs have been performed.
For each run, the number of evaluations needed to converge is recorded.
The maximum of this number over the $30$ is the value of $k$ (given in Table \ref{evalMaxNC}).
For the F2NS and VEGAS,
we consider that the search converged when $S$ cover the whole NN, 
and no portal is available on this NN (local optimum plateau). 
The VEGAS algorithm has a single problem-independent parameter: $C$,
which allows to control the trade-off between the exploration and the exploitation of solutions neighborhood from the NN.
Following \cite{Fialho10}, multiple $C$-values are investigated: $C \in \{ 0.0001, 0.001, 0.01, 0.1, 1, 10, 100, 500 \}$.
Let VEGAS$_n$ denote a VEGAS instance with $C=n$.

\begin{table}[htb]
\caption{Maximum number of moves $k$ on the same NN for NC.}
\label{evalMaxNC}
\begin{center}
\begin{tabular}{c|cccc}
 q $\setminus$ K  & 2 & 4 & 6 & 8\\
\hline
 2 & 23,772 & 27,950 & 7,733 & 6,143 \\
 3  &  1,891 & 1,648 & 1,987 & 1,921 \\
 4  &  8,198 & 2,000 & 3593 & 1189 \\
\end{tabular}
\end{center}
\end{table}

\begin{figure*}[t]
\begin{center}
\begin{tabular}{cc}
\includegraphics[scale=0.53]{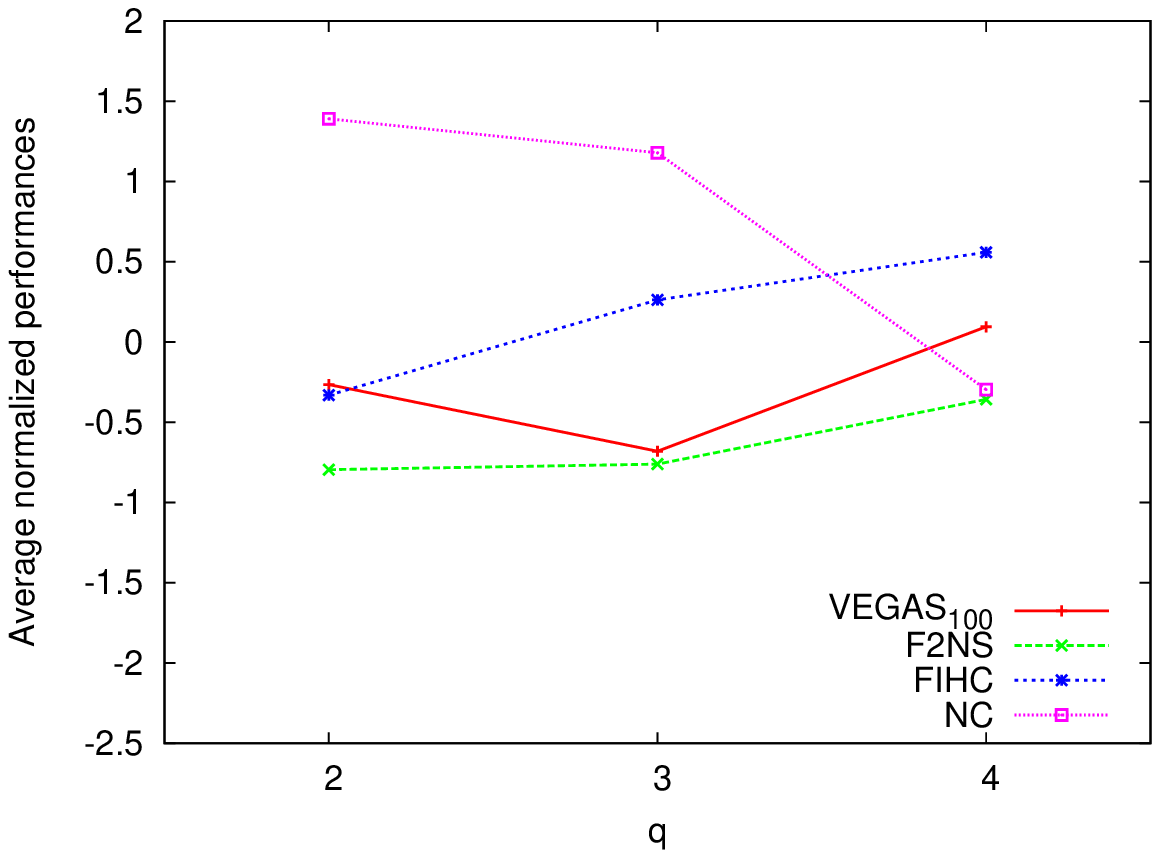} & \includegraphics[scale=0.53]{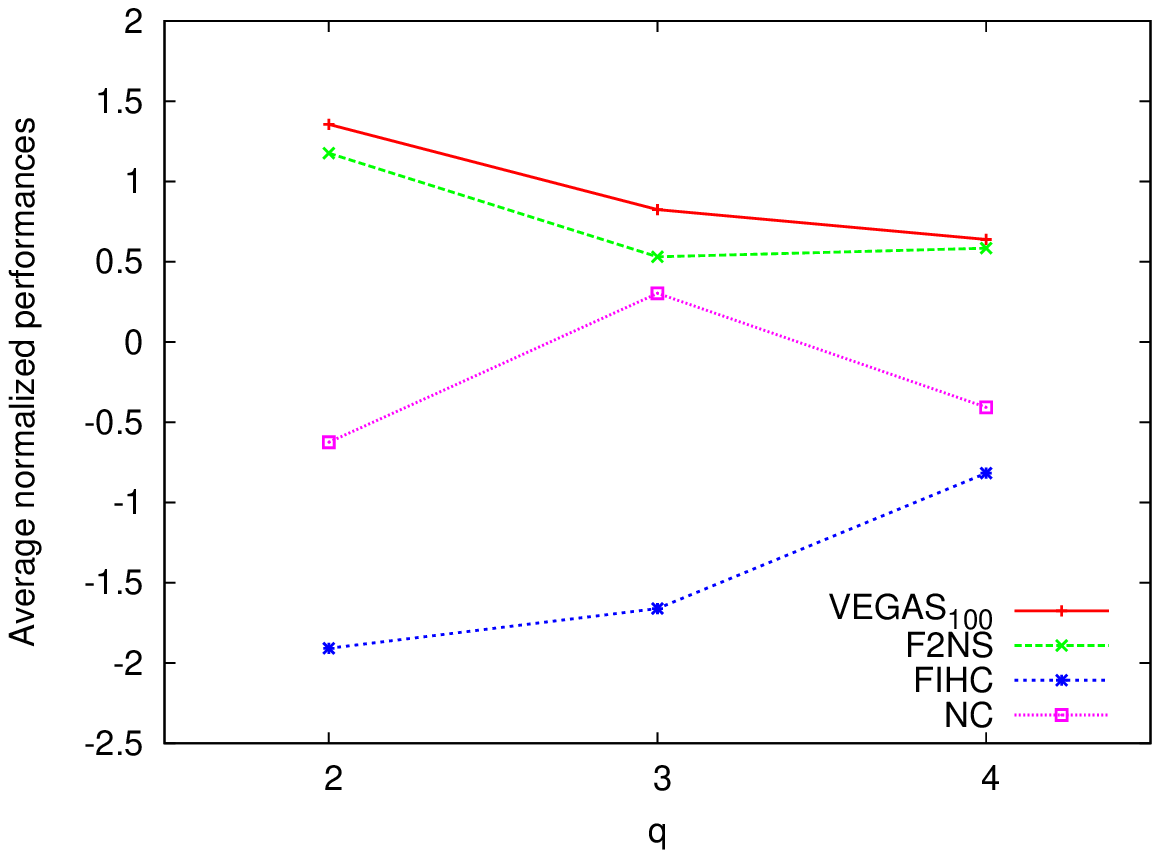} \\
(a) & (b) \\
\includegraphics[scale=0.53]{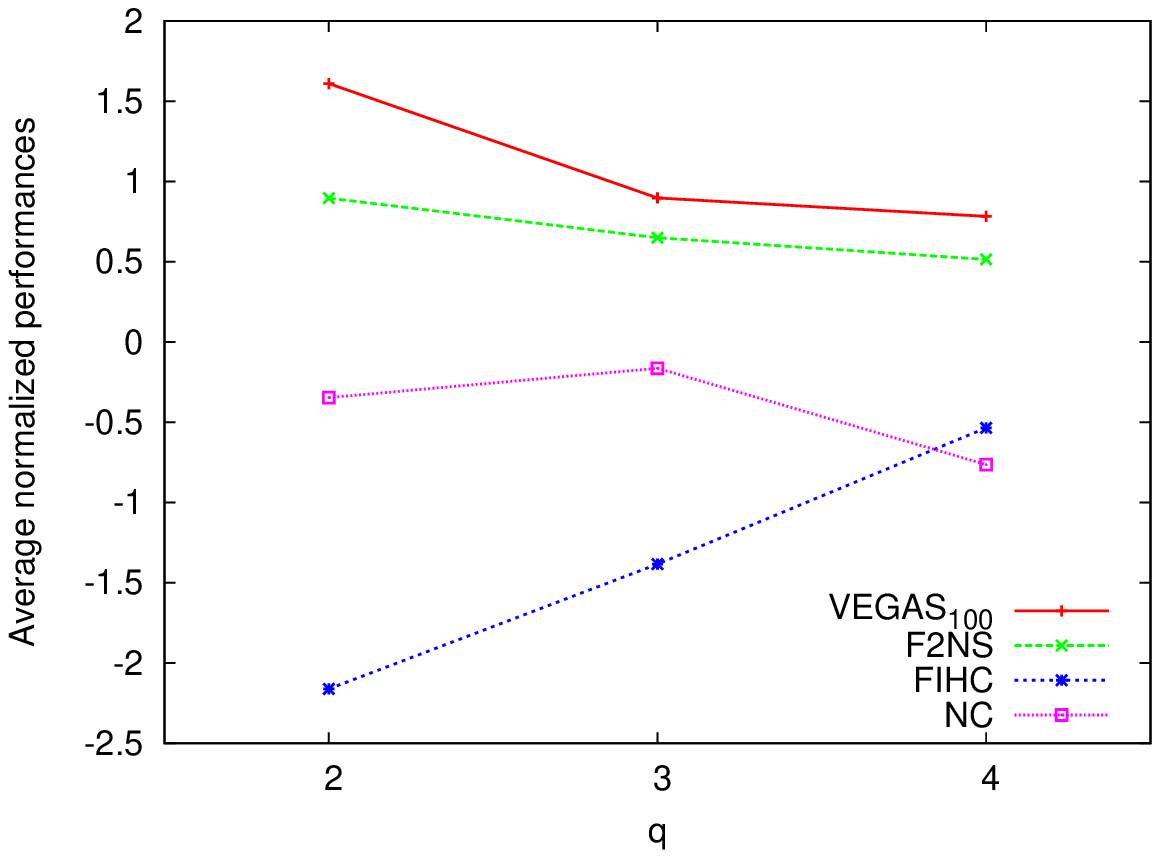} & \includegraphics[scale=0.53]{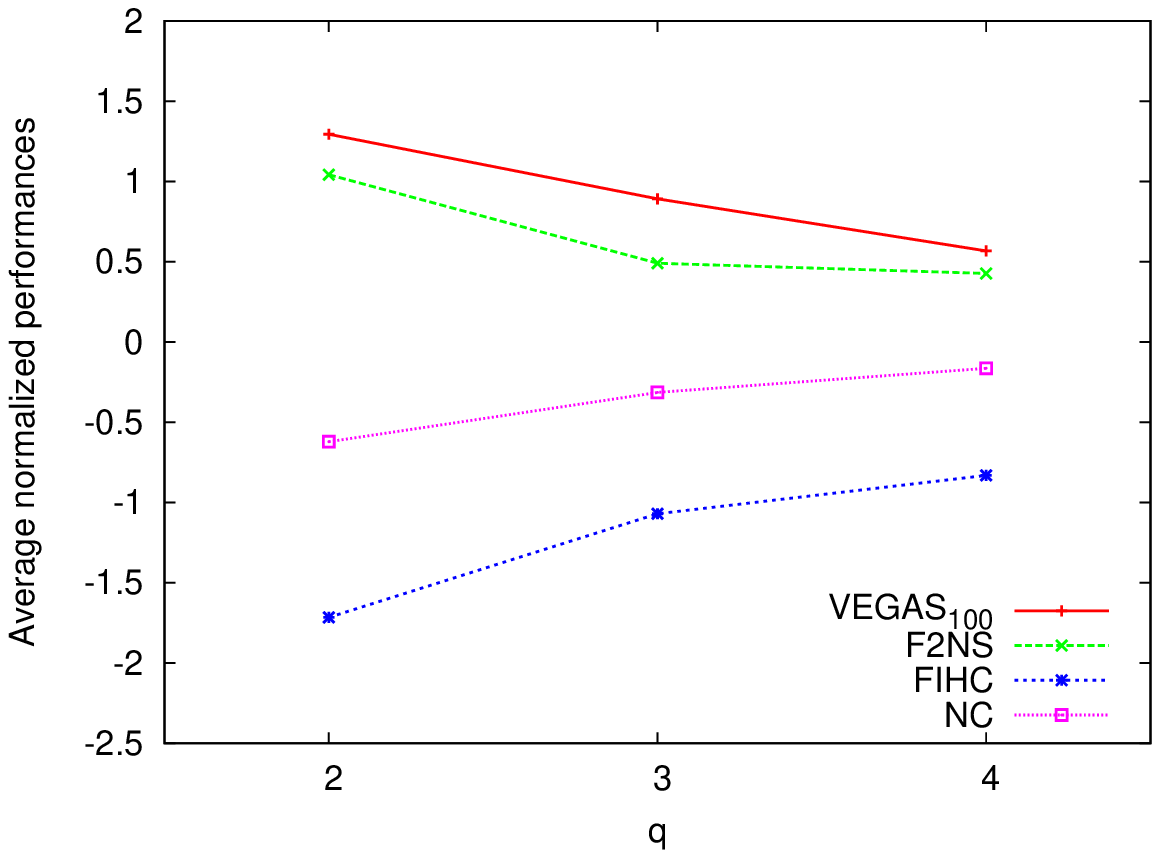} \\
(c) & (d) \\
\end{tabular}
\caption{Average normalized fitness found according to parameter $q$ after $10^5$ evaluations for
(a) $K=2$, (b) $K=4$, (c) $K=6$, (d) $K=8$. Results for VEGAS$_{100}$, F2NS, FIHC and NC.}
\label{perfAlgos}
\end{center}
\end{figure*}
%

%-------------------------------------
\subsection{Experimental Results and Discussion}

The following experiments first deal with the overall performance of the VEGAS algorithm against F2NS, FIHC and NC.
Next, the influence of the parameter~$C$ is deeply analyzed and 
first conclusions on the link between $C$-values and the dynamics of the algorithm are given.

%-------------------------------------
\subsubsection{Performance Analysis}

In this section, the four algorithms under study are compared with each other. 
As far as VEGAS is concerned, only one parameter ($C$) is to be set. 
The influence of this parameter is studied in the next section. 
Here we use a value of $C=100$, as it leads to overall good performance.

With respect to the non-linearity ($K$) and to the level of neutrality ($q$),
the fitness values of the solutions found are not comparable.
Hence, in order to compare the performance of the four algorithms according to the parameters $K$ and $q$,
the fitness values have been normalized using the average and the standard deviation of all fitness values found for the same problem instance.
Such an approach brings the average performance around zero and enlarges the extreme behaviors.
The average value $\bar{f}$ and the standard deviation $\sigma_f$ of all fitness values are computed with the $100$ runs performed by each algorithm. 
For a fitness value~$f$, its normalized value $\tilde{f}$ is set to: $\frac{f-\bar{f}}{\sigma_f}$. 
The performance of a given algorithm on a particular $NK_q$-landscape is given in terms
of this average $\tilde{f}$-value over the $100$ executions.

The respective performance of all the algorithms is given in Figure \ref{perfAlgos}.
For each $K$-value, the performance is plotted with respect to $q$.
For $K \in \{4, 6, 8\}$, VEGAS$_{100}$ and F2NS clearly outperform FIHC and NC. 
The comparison of the performance of VEGAS and F2NS is more difficult as, even if VEGAS always outperforms F2NS,
both are very close and the difference may not be statistically significant (see below). 
%A Wilcoxon test will be used to analyze this difference.
For $K=2$, Figure \ref{perfAlgos} (a) does not show a clear result.
Indeed, for $q=\{2,3\}$, NC gives the best performance,
but it becomes almost the worst approach for $q=4$, where the best approach seems to be FIHC.

In order to assess these results,
a Wilcoxon two-sample paired signed rank test is used with the null hypothesis that the median performance of the paired differences of the two algorithms under comparison is null.
Table \ref{algoWilcoxon} gives the output of the Wilcoxon test.
%
%%Table Wilcoxon

\begin{table*}
\begin{scriptsize}
\begin{center}
\caption{Wilcoxon paired tests on the 100 runs between the 4 algorithms. $=$ means both algorithms are not significantly different, $>$ means the algorithm of the row outperforms the one of the column and $<$ means for the contrary.}
\label{algoWilcoxon}
\begin{tabular}{cc|cccp{0.1cm}cccp{0.1cm}ccc}
\multicolumn{2}{c}{} & \multicolumn{3}{c}{q=2} & &  \multicolumn{3}{c}{q=3} & &  \multicolumn{3}{c}{q=4} \\
\cline{3-5} \cline{7-9} \cline{11-13}\\
\multicolumn{2}{c}{} & VEGAS & F2NS & FIHC & & VEGAS & F2NS & FIHC &  & VEGAS & F2NS & FIHC  \\
\hline
\multicolumn{13}{l}{}\\

$K=2$  & F2NS & $=$ &     &     &   & $=$ &     &     &  & $<$  \\
	   & FIHC & $=$ & $=$ &     &   & $<$ & $<$ &     &  & $>$ & $>$ \\
	   &  NC  & $>$ & $>$ & $>$ &   & $>$ & $>$ & $>$ &  & $<$ & $=$ & $<$ \\
\multicolumn{13}{l}{}\\

$K=4$  & F2NS & $=$ &     &     &   & $<$ &     &     &  & $=$   \\
	   & FIHC & $<$ & $<$ &     &   & $<$ & $<$ & $-$ &  & $<$ & $<$ \\
	   &  NC  & $<$ & $<$ & $>$ &   & $<$ & $<$ & $>$ &  & $<$ & $<$ & $>$ \\
\multicolumn{13}{l}{}\\

$K=6$  & F2NS & $<$ &     &     &   & $=$ &     &     &  & $=$ &  \\
	   & FIHC & $<$ & $<$ &     &   & $<$ & $<$ &     &  & $<$ & $<$ \\
	   &  NC  & $<$ & $<$ & $>$ &   & $<$ & $<$ & $>$ &  & $<$ & $<$ & $=$ \\
\multicolumn{13}{l}{}\\
$K=8$  & F2NS & $=$ &     &     &   & $<$ &     &     &  & $=$    \\
	   & FIHC & $<$ & $<$ &     &   & $<$ & $<$ &     &  & $<$ & $<$ \\
	   &  NC  & $<$ & $<$ & $>$ &   & $<$ & $<$ & $>$ &  & $<$ & $<$ & $>$ \\
\end{tabular}
\end{center}
\end{scriptsize}
\end{table*}

\begin{figure*}[t]
\begin{center}
\begin{tabular}{ccc}
\includegraphics[scale=0.40]{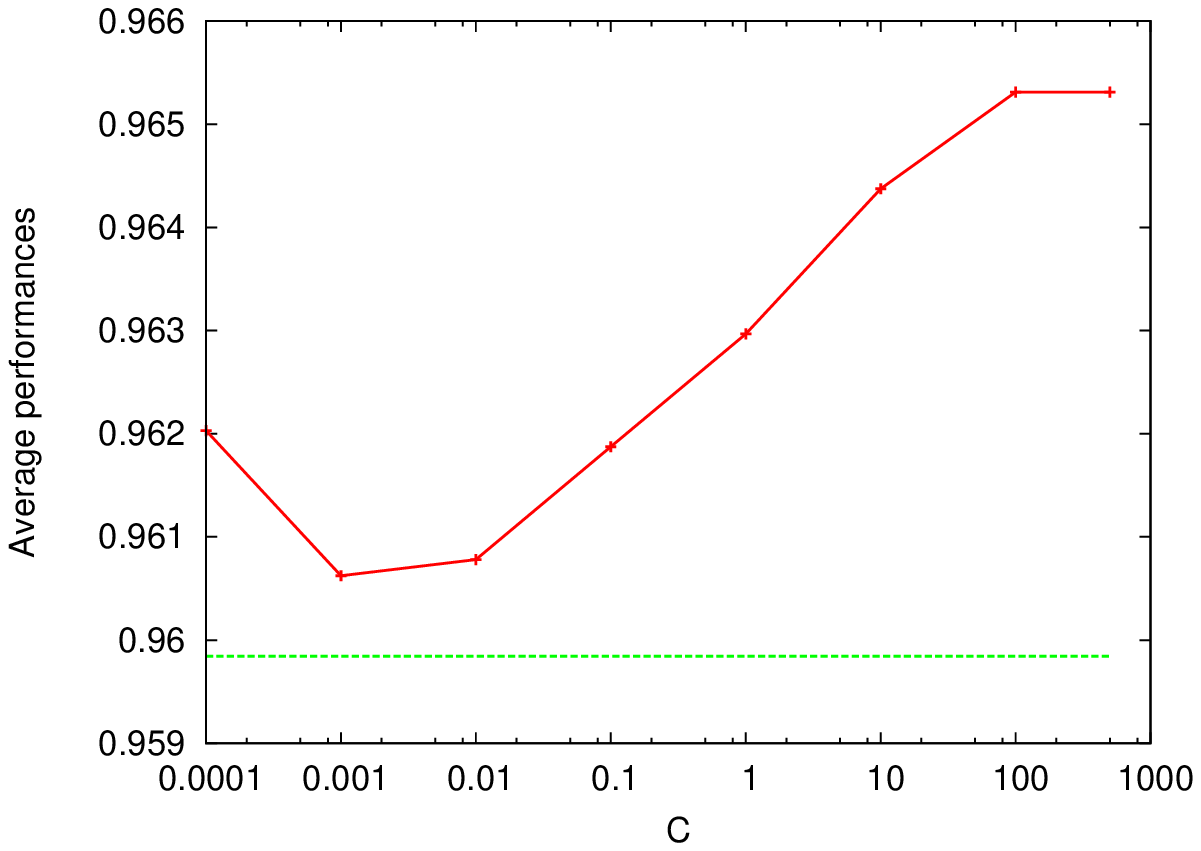} & \includegraphics[scale=0.40]{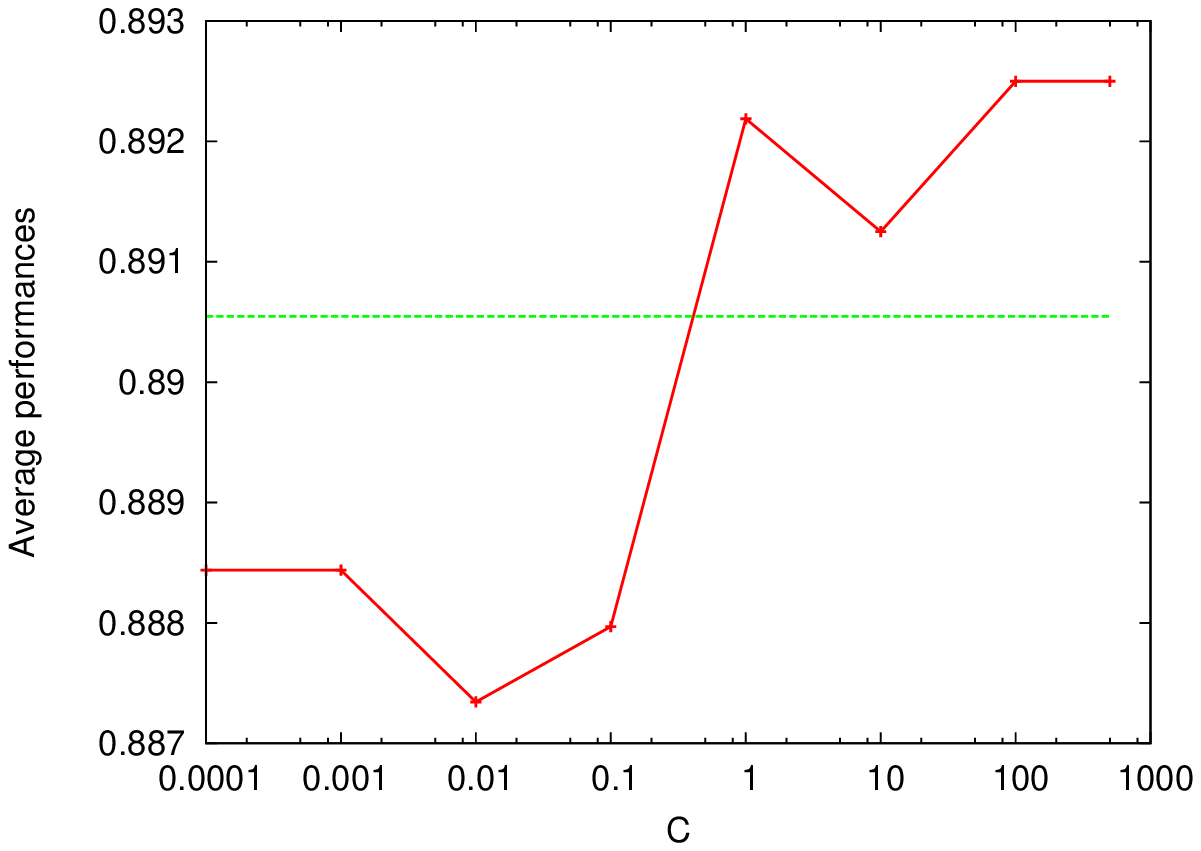} &\includegraphics[scale=0.40]{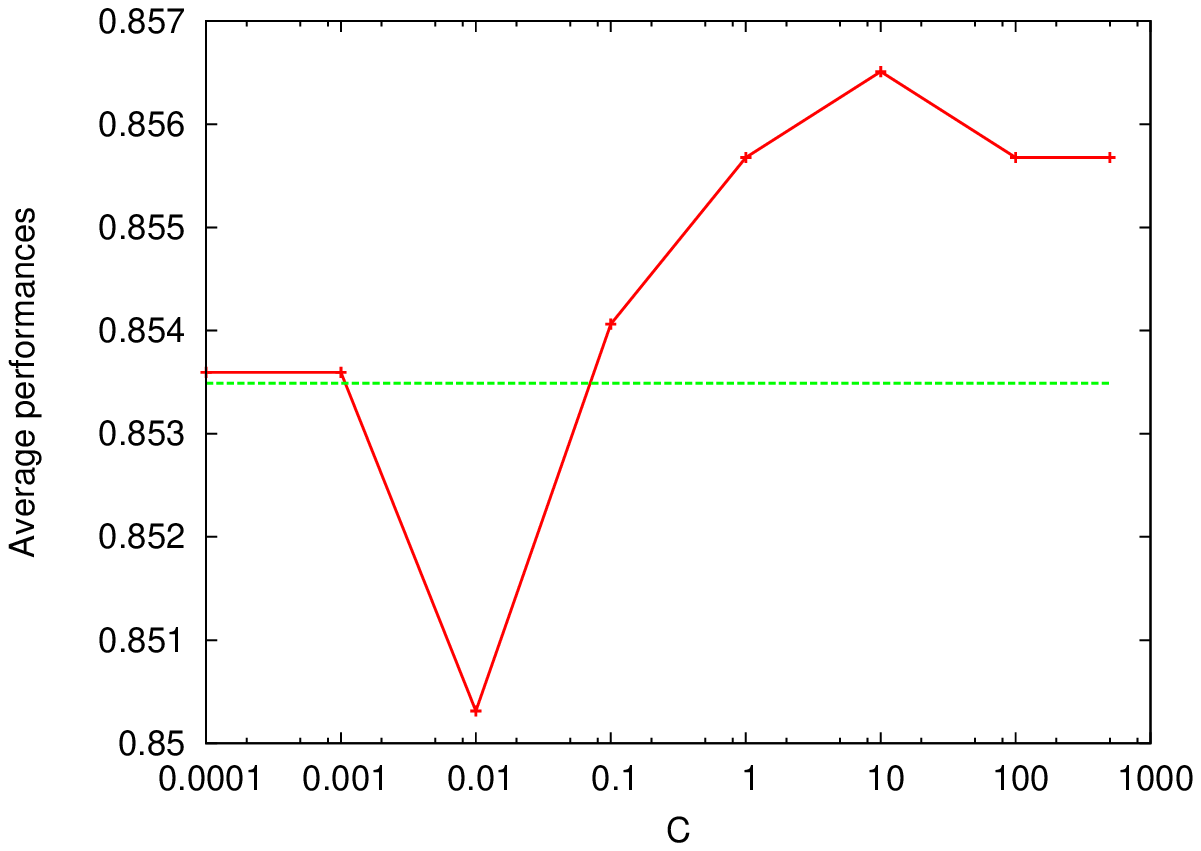}  \\ 
(a) & (b) & (c) \\
\end{tabular}
\caption{Average fitness found by VEGAS$_C$ as a function of C.  $K=6$ (a) $q=2$ (b) $q=3$ (c) $q=4$. The horizontal dotted line is the performance of F2NS.}
\label{perfVF}
\end{center}
\end{figure*} 

These results confirm that:
\begin{itemize}
\item Except for $K=2$, VEGAS and F2NS always significantly outperform FIHC and NC. 
It shows that ($i$)~it is worth exploring the NN (in contrast to FIHC), and that ($ii$) the way to do it has a large influence. 
Indeed, even if a method selects a solution at random on the NN (like F2NS),
it may outperform a method that always selects the last-evaluated solution (like NC).
\item VEGAS is never outperformed by F2NS. 
It shows that guiding the search over a NN allows to obtain better solutions.
\end{itemize}

In summary, exploring NN is a good way to guide the search over neutral landscapes,
at last for a reasonable level of neutrality.
However, this must be done carefully, and it seems to be better to randomly choose the next solution to explore than making always the same arbitrary choice. 
An alternative that shows interesting results is to pursue the search from the solution with the best evolvability.

%-------------------------------------
\subsubsection{Impact of Parameter $C$}

In this section, we analyze the performance of VEGAS according to the setting of its (single) problem-independent parameter: $C$.
Comparing the results obtained for all the instances, the efficiency of VEGAS does not seem to be clearly sensitive to the parameter $C$. 
This is confirmed by the Wilcoxon paired test that indicates that no general trend can be found. 
However, for some instances, VEGAS with exploration ($C > 1$) outperforms VEGAS with exploitation ($C < 1$) in average.

This is illustrated in Figure \ref{perfVF} for $K=6$, 
that gives the average fitness values obtained according to parameter $C$. 
Similar figures were obtained for $K=4$ and $K=8$. 
Indeed, the exploration of NN~($C > 1$) gives better results than their exploitation ($C < 1$).
In Figure \ref{perfVF}, there is also a dotted line representing the average fitness value obtained with F2NS
(when a solution is chosen at random among the NN). 
This confirms that VEGAS$_{100}$ outperforms F2NS,
but also that both methods obtain good performance,
F2NS may produce better results than some versions of VEGAS, typically when ${C<1}$.

%-------------------------------------
\subsubsection{Impact of Neutrality}

\begin{figure}[t]
\begin{center}
\includegraphics[scale=0.53]{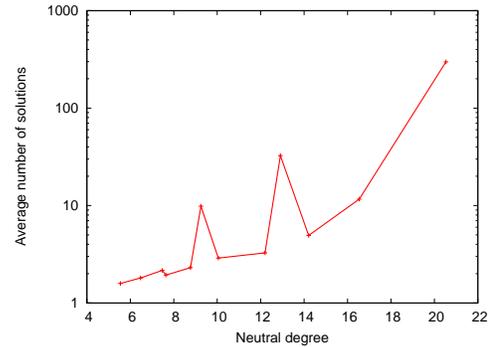}
\caption{Average number of evaluated solutions per NN with respect to the neutral degree.}.
\label{sizeNN_degN}
\end{center}
\end{figure}

%% trop moche
\begin{table}%[h]
\caption{Average neutral degree of the NKq instances.}% used for the experiments.}
\label{tabDegN}
\begin{center}
\begin{tabular}{c|cccc}
 q $\setminus$ K  & 2 & 4 & 6 & 8\\
\hline
2 & 20.53 & 16.54 & 14.21 & 12.2 \\
3  &  12.91 & 10.05 & 8.77 & 7.64 \\
4  &  9.25 & 7.47 & 6.47 & 5.54 \\
\end{tabular}
\end{center}
\end{table}

Table \ref{tabDegN} gives the average value, over 10 000 random solutions,
of the neutral degree according to the $NK_q$-landscape under study.
The neutrality decreases when $K$ and/or $q$ increase.
Figure \ref{sizeNN_degN} gives the average number of solutions evaluated per NN by VEGAS$_{100}$ %on each NN
according to the neutral degree.
It shows the influence of the neutrality on the number of evaluated solutions on each encountered NN.
This number increases exponentially with the neutral degree. 
Let us indicate that the two picks correspond to $K=2, q=\{3,4\}$. 
As we already pointed out, when $K=2$ (small epistasis), VEGAS  has a different behavior.

%-------------------------------------
\subsubsection{Exploration \textit{vs.} Exploitation}

\begin{figure*}[t]
\begin{center}
\begin{tabular}{cc}
\includegraphics[scale=0.53]{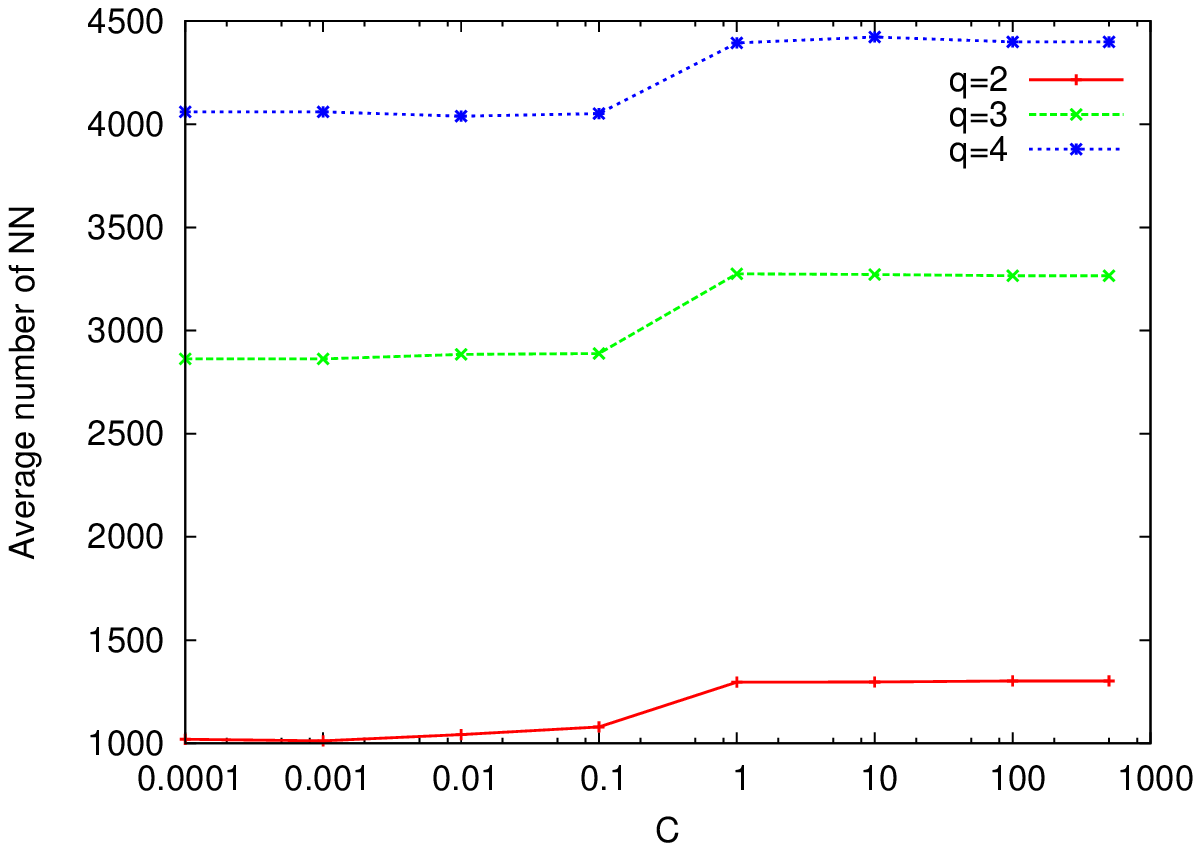} & \includegraphics[scale=0.53]{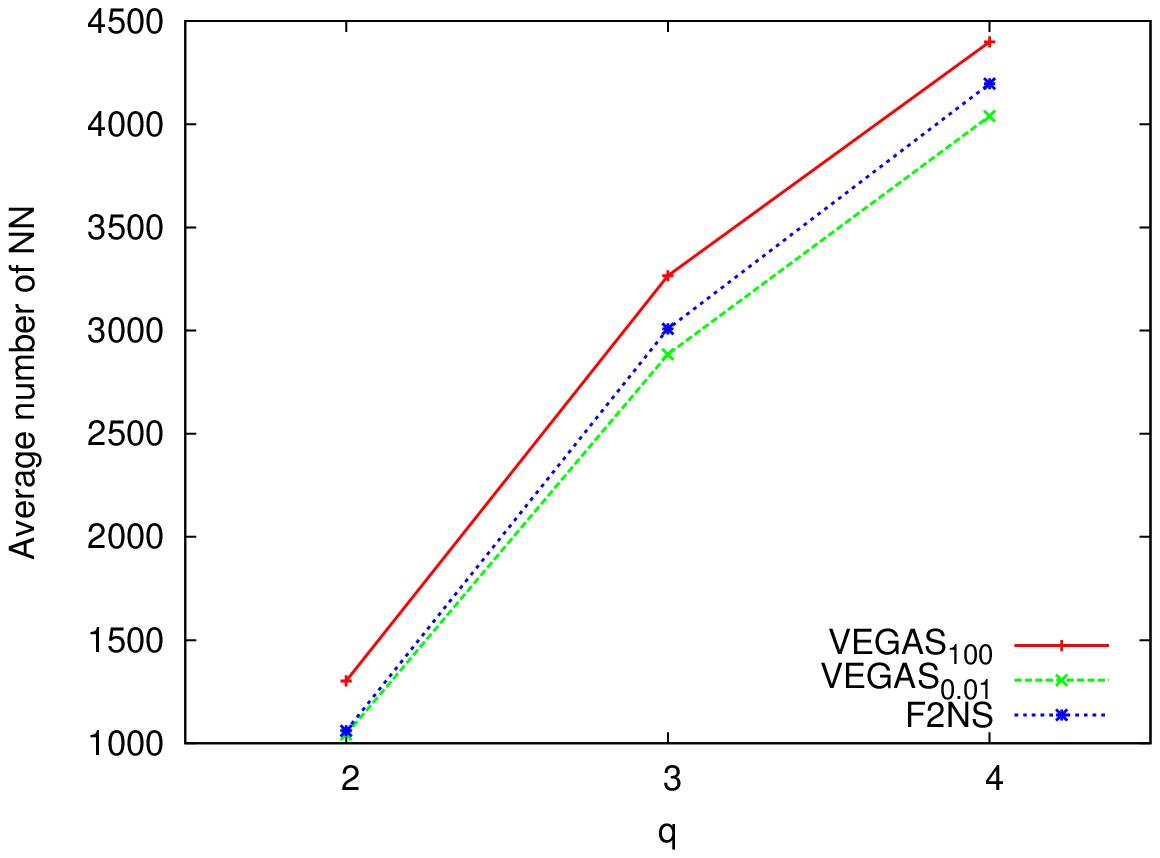} \\
(a) & (b) \\
\includegraphics[scale=0.53]{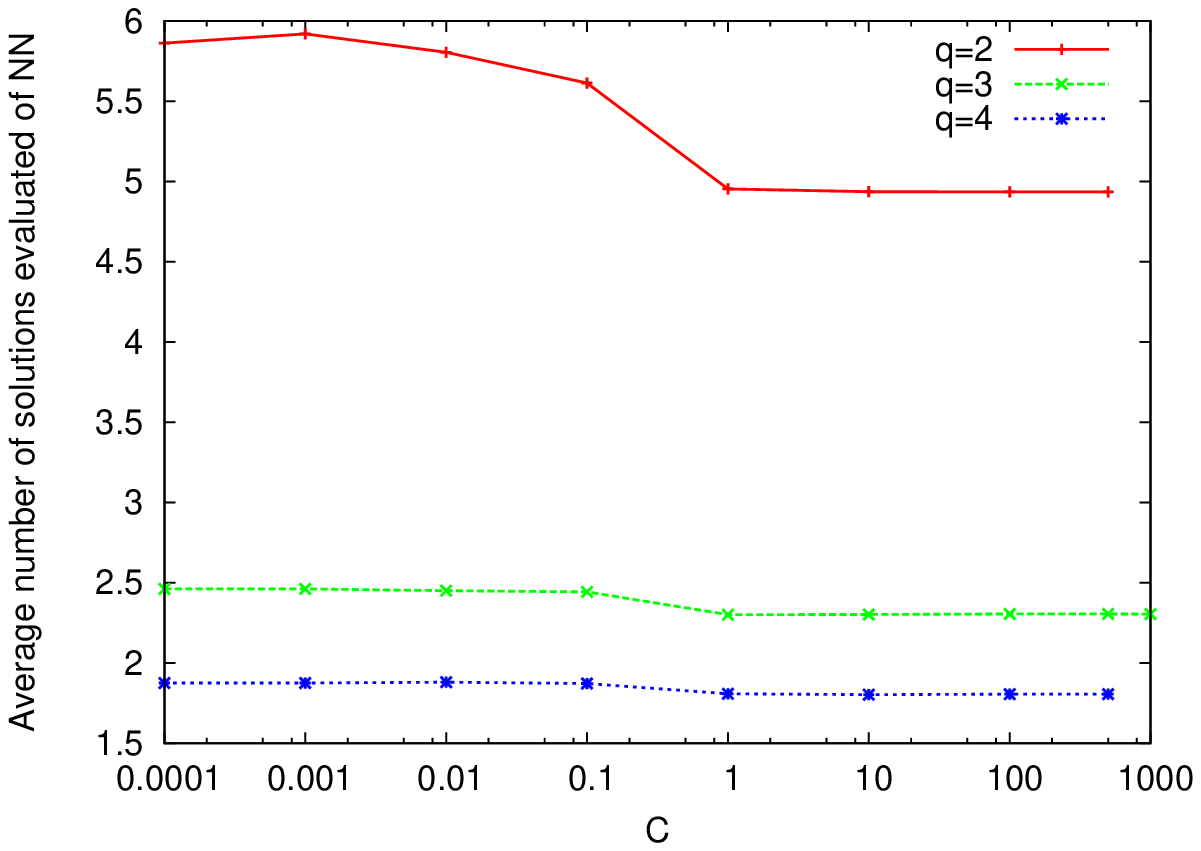} & \includegraphics[scale=0.53]{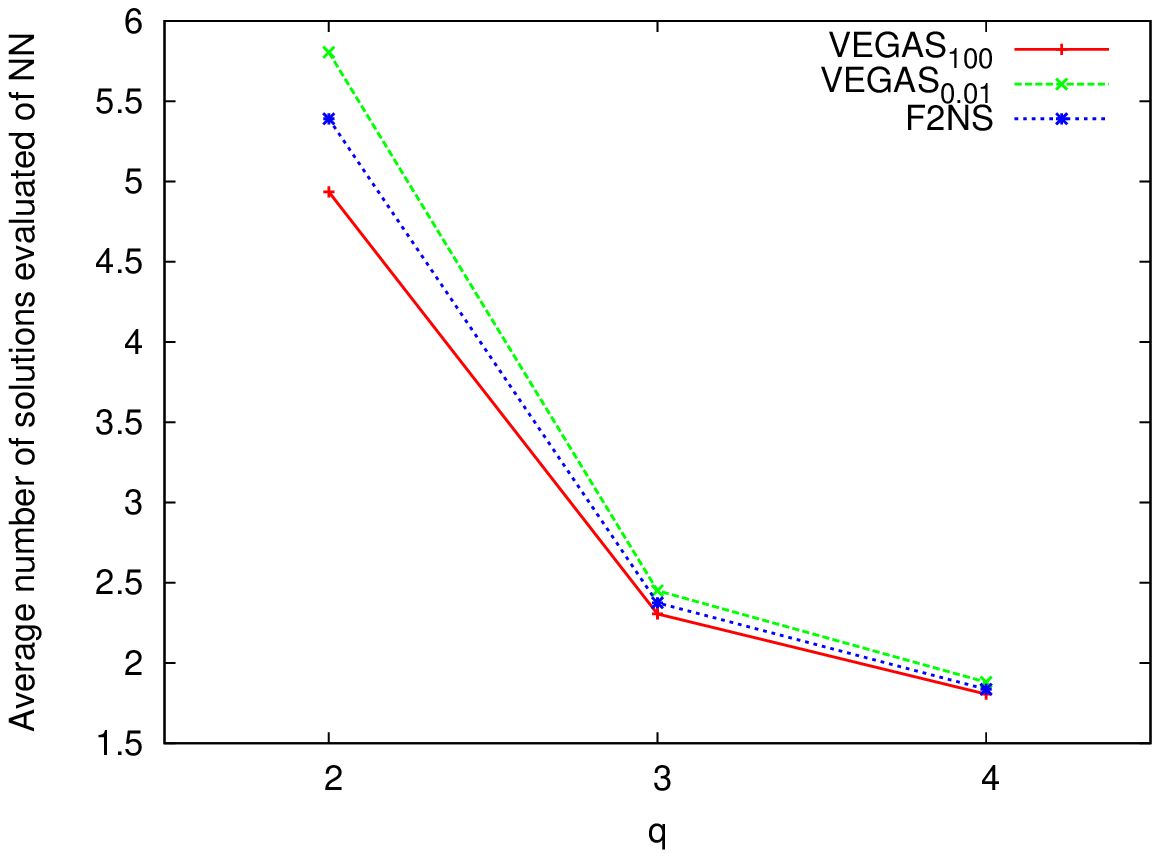} \\
(c) & (d) 
\end{tabular}
\caption{The average number of NN (top) and the average number of evaluated solutions of each NN (bottom). On the left, the average quantities are compared according to $C$ according to the parameter $q$. On the right, they are compared according to $q$ between 2 VEGAS algorithms (with $C=\{100,0.01\}$) and F2NS. for all plots, $K=6$.}
\label{NeutralNetworks}
\end{center}
\end{figure*}

Previous experiments show that the exploration of a larger number of solutions from the NN gives, in general, better performance. 
In order to analyze this in more details, some statistics have been computed to study the dynamics of the search. 
Two main statistics are computed here:
($i$) the number of NN evaluated,
($ii$) the number of solutions evaluated on each NN, which corresponds to the size of the neutral networks explored part.

Figure \ref{NeutralNetworks}
gives (a) the average number of NN and (c) the average number of solutions evaluated on each NN according to~$C$.
First, there is a clear difference between VEGAS with exploitation and VEGAS with exploration. 
Indeed, for $C \geq 1$, average values are similar.
The same happens for $C < 1$.
The different curves may be cut around $C=0.5$ into two homogeneous parts.
Second, the higher the average number of NN, the smaller the average number of evaluated solutions \textit{per} NN.
This attests a large difference on the behavior of the VEGAS algorithm regarding the $C$-values. 
In other words, when $C$ is turned to exploration, more NN are explored, but the portion of evaluated solutions is small.
When $C$ is turned to exploitation, few NN are explored, but they are deeply evaluated.
Such dynamics may explain that VEGAS with exploration gives, in some cases, a better performance than VEGAS with exploitation.

Figure \ref{NeutralNetworks} (b) and (d) also gives the same values with respect to $q$
for different algorithms: VEGAS with exploration (VEGAS$_{100}$), VEGAS  with exploitation (VEGAS$_{0.01}$) and F2NS (random choice).
Let us remark that similar trends happen for $K \in \{4,8\}$.
It appears that the F2NS approach has a behavior ``in-between'' the two VEGAS algorithms.  
As we previously seen on Figure \ref{perfVF}, the performance of F2NS is either below VEGAS or ``in-between'' the two VEGAS variants. 
A natural conclusion is that this balance between exploration and exploitation is a critical issue for the performance of the  algorithm.

%============================================================================
\section{Conclusions and Future Works}

This work proposes a new methodology to deal with neutral combinatorial optimization problems.
In this approach, all solutions identified on a plateau are considered in order to help the search to progress.
Then, the most promising solution evaluated on the plateau is selected adaptively, based on the evolvability of solutions.
VEGAS is an adaptive search algorithm using the multi-armed bandit framework
and the `area under the curve' credit assignment principle \cite{Fialho10}.

An experimental study on $NK$-landscapes with neutrality has been conducted.
It first shows that randomly choosing a solution on the plateau outperforms a netcrawler-based multi-start local search for a reasonable level of neutrality.
The experimental analysis also shows that VEGAS generally gives better results than selecting a solution at random on the plateau.
The VEGAS dynamics is different depending on the level of neutrality.
Moreover, VEGAS requires a single problem-independent parameter, that allows to tune the trade-off between the exploration and the exploitation of the plateau.
The influence of this parameter on the search dynamics has been deeply analyzed.

This approach shows encouraging results and open future research directions.
As in adaptive operator selection \cite{Fialho10}, we need to test others credit assignment methods, probably more specific to neutral landscapes.
Moreover, similar experiments will allow to better understand the dynamics of the VEGAS algorithm
on other combinatorial optimization problems where neutrality arises, such as in flowshop scheduling.

\bibliographystyle{abbrv}
%\bibliography{Marmion,NKbasins,llncs}

\end{document}